\documentclass[lettersize,journal]{IEEEtran}
\usepackage{amsmath,amsfonts,amssymb}
\usepackage{algorithmic}
\usepackage{algorithm}
\usepackage{array}
\usepackage[caption=false,font=normalsize,labelfont=sf,textfont=sf]{subfig}
\usepackage{textcomp}
\usepackage{stfloats}
\usepackage{url}
\usepackage{verbatim}
\usepackage{graphicx}
\usepackage{cite}
\usepackage{booktabs}
\usepackage[table]{xcolor}
\usepackage{multirow}
\usepackage{cleveref}
\crefname{figure}{Fig.}{Figs.}
\Crefname{figure}{Fig.}{Figs.}
\crefname{table}{Table}{Tables}
\Crefname{table}{Table}{Tables}
\crefname{section}{Section}{Sections}
\Crefname{section}{Section}{Sections}
\crefname{equation}{Eq.}{Eqs.}
\Crefname{equation}{Eq.}{Eqs.}
\newcommand{\eg}{e.g.\@}

\begin{document}

\title{Spa3R: Predictive Spatial Field Modeling for 3D Visual Reasoning}

\author{Haoyi Jiang, Liu Liu, Xinjie Wang, Yonghao He, Wei Sui, Zhizhong Su, Wenyu Liu,~\IEEEmembership{Senior Member,~IEEE}, Xinggang Wang,~\IEEEmembership{Senior Member,~IEEE}
\thanks{This work was supported by the National Natural Science Foundation of China under Grant 62376102.}
\thanks{Corresponding author: Xinggang Wang (xgwang@hust.edu.cn).}
\thanks{Haoyi Jiang, Wenyu Liu, and Xinggang Wang are with the School of Electronic Information and Communications, Huazhong University of Science and Technology, Wuhan 430074, China.}
\thanks{Liu Liu, Xinjie Wang, and Zhizhong Su are with Horizon Robotics, Beijing 100094, China.}
\thanks{Yonghao He and Wei Sui are with D-Robotics, Shenzhen 518055, China.}}

\maketitle

\begin{abstract}
    Vision-language models excel at 2D visual understanding but remain limited in 3D spatial reasoning. Existing approaches either depend on explicit 3D modalities, which limits scalability, or inject partial, view-conditioned geometric priors and leave the language model to recover global scene structure from sparse cues. We introduce \textbf{Spa3R}, a self-supervised framework that learns a unified, view-invariant spatial representation from unposed multi-view RGB images. Its Predictive Spatial Field Modeling objective compresses context views into a compact latent representation and predicts aligned geometric and semantic feature fields at novel viewpoints, thereby encouraging coherent encoding of scene geometry and layout. We integrate the pre-trained Spa3R Encoder into a vision-language model through a lightweight residual cross-attention adapter, yielding \textbf{Spa3-VLM} and grounding language reasoning in global spatial context. Spa3-VLM achieves an average score of 58.6\% on VSI-Bench and delivers leading or competitive performance across three additional spatial reasoning benchmarks. These results demonstrate that predictive spatial representation learning provides an effective visual foundation for 3D reasoning.
\end{abstract}


\section{Introduction}

\IEEEPARstart{T}{he} ability to perceive and reason about the 3D world is fundamental to spatial intelligence, underpinning applications from autonomous navigation to robotic manipulation. Despite remarkable progress in 2D image understanding, vision-language models (VLMs) remain limited in modeling 3D geometry and spatial relations~\cite{VSIBench}. This limitation partly stems from pre-training objectives that provide little incentive to integrate multiple views into a coherent 3D scene representation.

To endow VLMs with 3D spatial awareness, a straightforward strategy is to train on large-scale multi-view data paired with spatial question-answering (QA) annotations. However, this approach requires extensive annotated data and asks the language model to acquire spatial understanding largely through linguistic supervision. An alternative line of work~\cite{3DLLM, LL3DA} incorporates explicit 3D modalities such as depth-derived or reconstructed point clouds. These methods are geometrically grounded, but their dependence on specialized sensors restricts real-world scalability and applicability. Consequently, recent advances in geometry foundation models~\cite{DUSt3R, VGGT} have inspired methods that augment VLMs with geometric priors extracted from multiple views~\cite{VGLLM, SpatialMLLM, VLM3R}. Yet, these methods typically provide the VLM with partial, view-conditioned features derived from a limited set of observations. The language model must therefore infer coherent scene structure from partial cues using only sparse, indirect supervision from spatial instruction tuning.

\begin{figure}[t]
    \centering
    \includegraphics[width=\columnwidth]{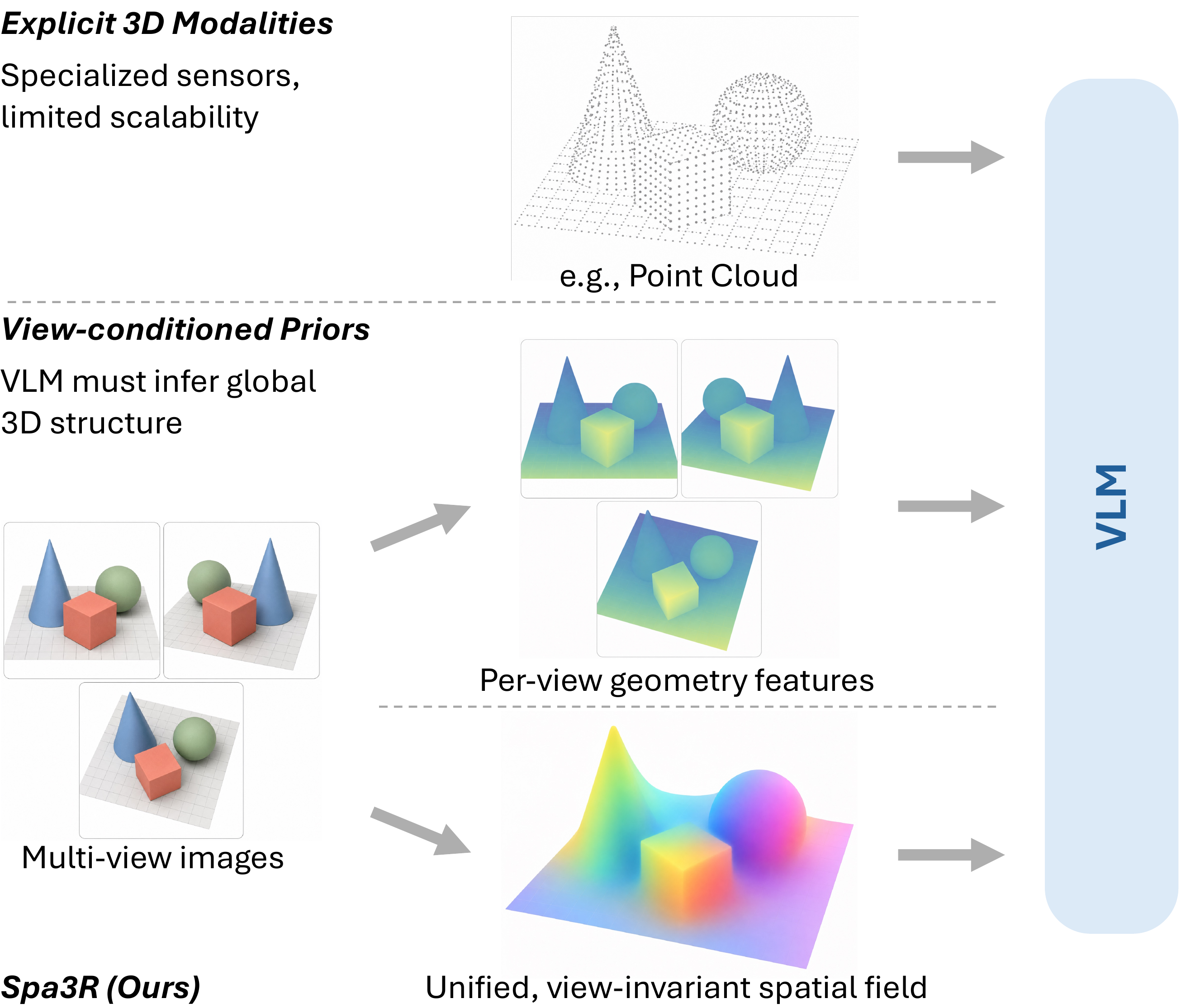}
    \caption{Comparison of paradigms for spatial reasoning. Explicit 3D
    modalities require specialized sensors, while view-conditioned priors
    leave the VLM to infer global 3D structure from per-view features. In
    contrast, Spa3R learns a unified, view-invariant spatial field directly
    from multi-view images for downstream VLM reasoning.}
    \label{fig:teaser}
\end{figure}

We instead explore whether spatial representations can be learned directly from visual observations through predictive modeling, without language supervision during representation pre-training. To this end, we introduce Spa3R, a self-supervised framework based on the Predictive Spatial Field Modeling (PSFM) paradigm. Given unposed context images, the Spa3R Encoder constructs a unified, view-invariant spatial representation of a 3D scene. Following the predictive principle of the Joint-Embedding Predictive Architecture (JEPA)~\cite{I-JEPA}, the decoder predicts spatial feature fields at queried novel views conditioned on this representation. PSFM creates an architectural bottleneck that encourages the encoder to capture scene geometry and layout that remain consistent across views, yielding a scene-level representation that extends beyond individual observations.

By decoupling spatial representation learning from linguistic reasoning, the pre-trained Spa3R Encoder can be used as a plug-in module for VLMs. We integrate it into Qwen2.5-VL~\cite{Qwen2_5VL} via a lightweight Residual Cross-Attention Adapter to form Spa3-VLM, allowing the VLM's native visual features to query the unified spatial context while retaining the base model's pre-trained knowledge. Experiments show that Spa3-VLM reaches an average score of 58.6\% on VSI-Bench~\cite{VSIBench} and consistently improves over the Qwen2.5-VL baseline across three additional spatial reasoning benchmarks.

Our contributions are threefold:

\begin{itemize}
    \item We identify a fundamental limitation of existing VLMs for spatial reasoning: the language model must infer coherent 3D scene structure from partial, view-conditioned features using only instruction-tuning supervision.
    \item We propose \textbf{Spa3R}, a self-supervised framework based on Predictive Spatial Field Modeling (PSFM). By predicting feature fields at novel views, Spa3R learns a unified, view-invariant spatial representation of scene geometry and layout.
    \item We present \textbf{Spa3-VLM}, which integrates the pre-trained Spa3R Encoder to ground VLM reasoning in global spatial context. Experiments show competitive results against existing spatial models, while controlled ablations demonstrate consistent gains over matched baselines.
\end{itemize}

\section{Related Work}

\subsection{3D Reconstruction and Scene Representation}

Recent 3D reconstruction methods have shifted from scene-specific optimization toward generalizable feed-forward inference. Neural Radiance Fields (NeRF)~\cite{NeRF} introduced implicit neural fields for scene representation, yet require costly per-scene training. Joint pose and radiance-field optimization, as in CT-NeRF~\cite{CTNeRF}, relaxes the calibrated-camera assumption but remains scene-specific and optimization-intensive. Subsequent methods, such as pixelNeRF~\cite{pixelNeRF} and MVSNeRF~\cite{MVSNeRF}, introduced conditional neural fields to learn generalizable 3D priors across multiple scenes. This feed-forward paradigm has recently been extended to 3D Gaussian Splatting~\cite{pixelSplat}, occupancy prediction~\cite{RenderOcc, SelfOcc, GaussTR}, and vision-based semantic scene completion~\cite{MixSSC}, enabling efficient 3D perception from images. Self-supervised implicit reconstruction has also been used to learn part-aware object representations~\cite{MaskedPaCONet}, highlighting reconstruction as a useful pretext task beyond photorealistic rendering.

In parallel, large-scale pre-training has fostered the development of geometry foundation models. The DUSt3R~\cite{DUSt3R, MASt3R} and VGGT~\cite{VGGT, PI^3} families have demonstrated robust 3D reconstruction by jointly estimating diverse 3D attributes, including point maps, depth, and camera parameters, from unposed images. These foundation models provide spatially grounded priors, inspiring hybrid frameworks~\cite{NoPoSplat, LSM, Uni3R} that combine geometric predictions with rendering-based refinement for 3D reconstruction and semantic understanding.

Recent work has explored Transformer-only architectures for novel-view synthesis (NVS), including LVSM~\cite{LVSM} and its variants~\cite{RayZer, UP-LVSM}. Without explicit 3D inductive biases, these models use Transformer capacity and scaling to infer geometric structure from 2D data. Spa3R draws inspiration from this predictive paradigm but differs in its learning objective. While LVSM prioritizes pixel-level synthesis for high-fidelity reconstruction, PSFM targets representation learning by predicting spatially grounded feature fields. The resulting latent representation encodes scene geometry and semantic relationships in a unified, view-invariant space for downstream spatial reasoning.

\subsection{Vision-Language Models for 3D Reasoning}

While VLMs excel at 2D image-text alignment, extending them to the spatial domain remains challenging. Early approaches relied on explicit 3D modalities, either through object-centric representations~\cite{Chat3D, ChatScene} or by directly processing point clouds~\cite{3DLLM, LL3DA, LEO} to ground language in 3D space. Viewpoint-aware 3D visual grounding further reduces spatial ambiguity by predicting a pseudo-embodied viewpoint~\cite{PseudoEV}; this broader line of research is reviewed in the 3D dense-captioning literature~\cite{DenseCaption3DSurvey}. While these approaches benefit from strong geometric grounding, their dependence on specialized sensors (\eg, LiDAR) or preprocessed 3D data restricts their real-world scalability and applicability.

Recent work has expanded from 3D grounding to question answering and structured scene reasoning. Bias3D-VQA~\cite{Bias3DVQA} improves point-cloud-based 3D question answering by explicitly modeling and mitigating dataset biases. History-enhanced reasoning~\cite{HE3DSGR} incrementally constructs a consistent 3D semantic scene graph from RGB-D sequences, while spatial multimodal knowledge-driven prediction~\cite{SMK3DSGP} combines VLM-derived support relations with hierarchical visual and symbolic graphs. Although these methods enable structured 3D reasoning, they depend on explicit depth or point-cloud representations. To reason from more widely available 2D observations, GPT4Scene~\cite{GPT4Scene} reconstructs bird's-eye-view (BEV) representations from video features to capture spatiotemporal context, while approaches such as VG-LLM~\cite{VGLLM}, Spatial-MLLM~\cite{SpatialMLLM}, and VLM3R~\cite{VLM3R} augment VLMs with view-conditioned priors extracted from geometry foundation models. Meanwhile, models such as SenseNova-SI~\cite{SenseNova-SI} and VST~\cite{VST} enhance spatial awareness by fine-tuning general-purpose models on large-scale spatial datasets.

Nevertheless, these approaches typically expose the VLM to partial, view-conditioned geometric cues, leaving the language model to integrate them into a coherent 3D scene representation under sparse supervision. In contrast, PSFM learns a unified, view-invariant spatial representation through self-supervised predictive modeling, providing coherent scene-level context for spatial reasoning.

\section{Method}

\begin{figure*}[t]
    \includegraphics[width=\textwidth]{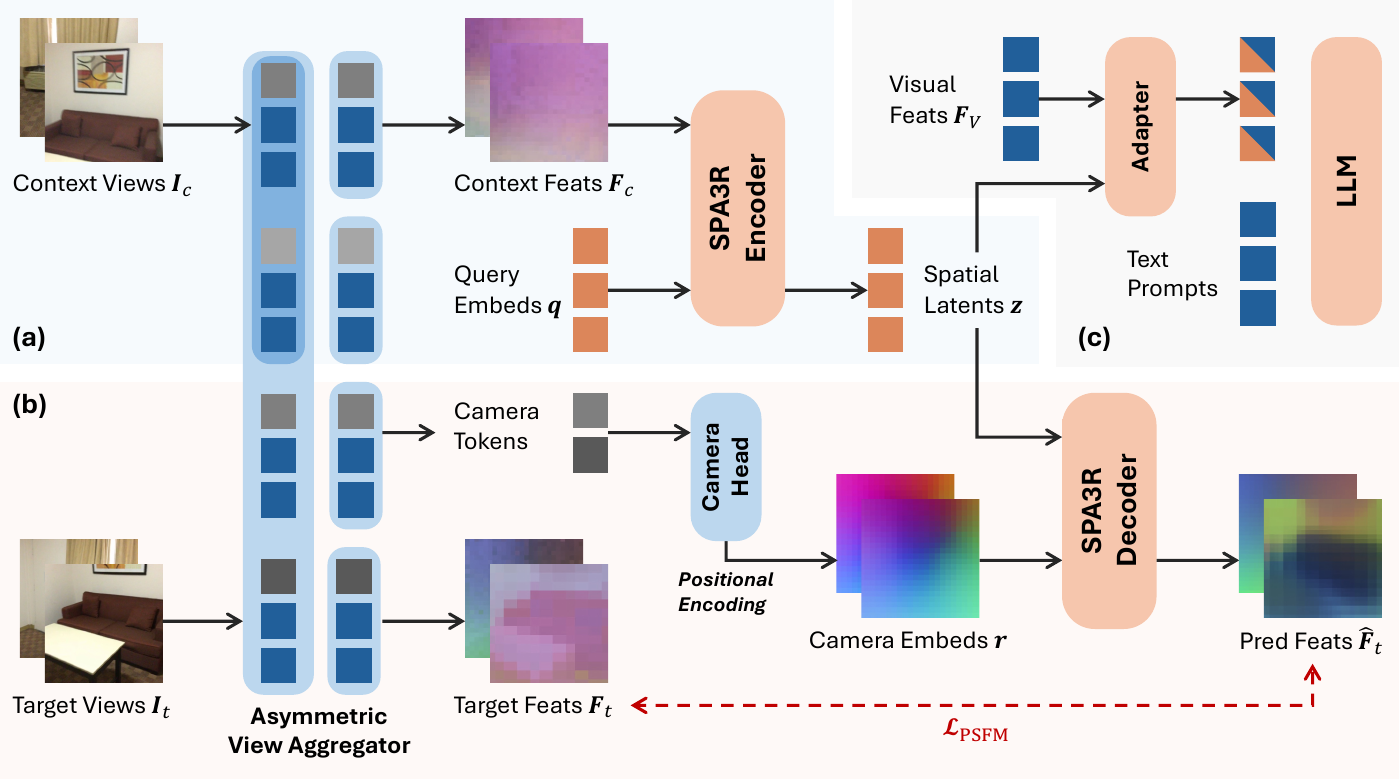}
    \caption{Overview of the Spa3R framework and Spa3-VLM integration. (a) The Spa3R Encoder maps unposed context views to a unified, view-invariant spatial latent representation $\boldsymbol{z}$. (b) The Spa3R Decoder synthesizes target features $\hat{\boldsymbol{F}}_t$ at queried novel views conditioned on $\boldsymbol{z}$ and target camera embeddings $\boldsymbol{r}$. (c) For downstream spatial reasoning, the pre-trained Spa3R Encoder produces $\boldsymbol{z}$, and a lightweight adapter fuses the VLM's native visual features $\boldsymbol{F}_V$ with the spatial latent through cross-attention.}
    \label{fig:framework}
\end{figure*}

We first formulate Predictive Spatial Field Modeling (PSFM) in \cref{sec:PSFM}, describe the Spa3R architecture in \cref{sec:Spa3R}, and explain its integration with vision-language models in \cref{sec:SPA_VLM}. \Cref{fig:framework} provides an overview.

\subsection{Predictive Spatial Field Modeling}
\label{sec:PSFM}

We formulate 3D spatial understanding as spatial field modeling. Specifically, we represent a 3D scene as a continuous feature field $f$ that maps each viewpoint in $\mathcal{V}$ to a view-centric feature map $\boldsymbol{F} \in \mathcal{F}$:
\begin{equation}
    f: \mathcal{V} \rightarrow \mathcal{F}.
\end{equation}
Each viewpoint is parameterized by an invertible lifted world-to-image projection matrix $\boldsymbol{P} \in \mathcal{V}$ constructed from the camera intrinsics and extrinsics.

The goal of PSFM is to infer a compact latent representation of scene geometry from a sparse set of $N_C$ context views $C = \{ (\boldsymbol{P}_c, \boldsymbol{F}_c) \}_{c=1}^{N_C}$. We formulate PSFM as a deterministic conditional neural process~\cite{ConditionalNP}: an amortized encoder $E_\phi$ maps the context set to a unified latent $\boldsymbol{z}$, and a decoder $D_\theta$ predicts the target feature map $\hat{\boldsymbol{F}}_t$ at a queried camera projection $\boldsymbol{P}_t$ conditioned on $\boldsymbol{z}$:
\begin{align}
    \boldsymbol{z}         & = E_\phi (C) = E_\phi \left( \{ (\boldsymbol{P}_c, \boldsymbol{F}_c) \}_{c=1}^{N_C} \right), \\
    \hat{\boldsymbol{F}}_t & = D_\theta (\boldsymbol{P}_t, \boldsymbol{z}).
\end{align}

During training, we sample a set of views $S$ from each scene and randomly partition it into a context set $C$ and a target set $T$. The learning objective minimizes the distance between the predicted features $\hat{\boldsymbol{F}}_t$ and their aligned targets $\boldsymbol{F}_t$:
\begin{equation}
    \mathcal{L}_{\text{PSFM}} = \mathbb{E}_{C, T \sim S} \left[ \sum_{t \in T} \text{dist} \left( D_\theta(\boldsymbol{P}_t, E_\phi(C)), \boldsymbol{F}_t \right) \right].
\end{equation}

This architectural bottleneck encourages $\boldsymbol{z}$ to retain view-invariant scene geometry and spatial relations rather than view-specific appearance.

\subsection{Learning a Unified Spatial Representation}
\label{sec:Spa3R}

Spa3R instantiates the PSFM paradigm with an encoder-decoder architecture comprising an Asymmetric View Aggregator, a Spa3R Encoder, and a Spa3R Decoder.

\paragraph{Asymmetric View Aggregator}
PSFM requires a canonical spatial feature field $f$ in which context features $\boldsymbol{F}_c$ and target features $\boldsymbol{F}_t$ are spatially aligned.
To construct this field, we adapt the pre-trained VGGT~\cite{VGGT} with an Asymmetric View Aggregator that extracts spatially aligned features through VGGT's global attention mechanism.
We employ an asymmetric attention mask in VGGT's global attention layers to prevent information leakage from target views to context views.
Given a batch of context views $C$ and target views $T$, we construct a view mask $\boldsymbol{M} \in \{0, -\infty\}^{L \times L}$ applied to the attention logits, where $L$ denotes the total sequence length. Specifically, tokens from target views may attend to all views ($C \cup T$), whereas tokens from context views may attend only to context views. For a query token belonging to view $i$ and a key token belonging to view $j$,
\begin{equation}
    \boldsymbol{M}_{ij} =
    \begin{cases}
        0       & \text{if } i \in T \text{ or } j \in C, \\
        -\infty & \text{otherwise}.
    \end{cases}
\end{equation}

This mechanism ensures that context features $\boldsymbol{F}_c$ are computed independently of the targets. Target queries can attend to all context and target views, so target-to-target attention contributes to the construction of the geometrically aligned targets $\boldsymbol{F}_t$. The frozen VGGT camera head predicts each target's intrinsics and extrinsics from the aggregated target tokens. The resulting target projection matrices $\boldsymbol{P}_t$ and geometric features $\boldsymbol{F}_t$ are therefore expressed in VGGT's reference coordinate system. Semantic targets are extracted independently from each target RGB image by the frozen DINOv3 encoder~\cite{DINOv3}.

\paragraph{Spa3R Encoder}
The Spa3R Encoder $E_\phi$ is a Transformer that compresses the context features $\boldsymbol{F}_c$ into a compact representation $\boldsymbol{z} \in \mathbb{R}^{N_q \times D}$. We initialize $N_q$ learnable query embeddings $\boldsymbol{q}$ and denote their output indices by $\mathcal{I}_q=\{1,\ldots,N_q\}$. The queries are concatenated with $\boldsymbol{F}_c$ and iteratively refined through Transformer layers, yielding the spatial representation $\boldsymbol{z}$:
\begin{align}
    \boldsymbol{H} & = \text{Transformer}(\text{Concat}[\boldsymbol{q}, \boldsymbol{F}_c]), \\
    \boldsymbol{z} & = \{\boldsymbol{H}_i : i \in \mathcal{I}_q\}.
\end{align}

\paragraph{Spa3R Decoder}
The Spa3R Decoder $D_\theta$ synthesizes the target features $\hat{\boldsymbol{F}}_t$ conditioned on the spatial latent $\boldsymbol{z}$ and the target projection matrix $\boldsymbol{P}_t$. The decoder uses two geometric components: ray-based querying and relative 3D positional encoding.
First, to encode the target-view frustum, we generate camera-space ray directions $\boldsymbol{d}$ for each homogeneous pixel coordinate $\tilde{\boldsymbol{u}}$ using the estimated camera intrinsics $\boldsymbol{K}^{\mathrm{cam}}_t$:
\begin{equation}
    \boldsymbol{d} = \operatorname{Normalize}\!\left((\boldsymbol{K}^{\mathrm{cam}}_t)^{-1} \tilde{\boldsymbol{u}}\right).
\end{equation}
These ray directions $\boldsymbol{d}$ are mapped to initial camera embeddings $\boldsymbol{r}$ via a linear projection.

Second, we use PRoPE~\cite{PRoPE} within the decoder to encode camera relationships at the attention level. Each camera-associated token $i$ is assigned an invertible lift $\boldsymbol{P}_i\in\mathbb{R}^{4\times4}$ of the world-to-image projection matrix formed from its predicted intrinsics and extrinsics, together with a two-dimensional patch coordinate in the corresponding image plane. Following PRoPE, these terms form a block encoding $\boldsymbol{D}^{\mathrm{PRoPE}}_i\in\mathbb{R}^{d_h\times d_h}$ for each attention head of dimension $d_h$; the projective $4\times4$ blocks are repeated to fill the designated channels and combined with two-dimensional RoPE blocks. For latent tokens without an associated camera, we use the identity PRoPE operator, corresponding to coordinate zero.

Let $\boldsymbol{q}_i$, $\boldsymbol{k}_j$, and $\boldsymbol{v}_j$ denote the attention query, key, and value vectors, respectively. The relative operator and attention output are
\begin{align}
    \boldsymbol{\mathcal{T}}_{i\leftarrow j}
    &= \boldsymbol{D}_{i}^{\mathrm{PRoPE}}
       (\boldsymbol{D}_{j}^{\mathrm{PRoPE}})^{-1}\in\mathbb{R}^{d_h\times d_h}, \\
    \alpha_{ij}
    &= \frac{\exp\!\left(\boldsymbol{q}_i^\mathsf{T}
       \boldsymbol{\mathcal{T}}_{i\leftarrow j}\boldsymbol{k}_j/\sqrt{d_h}\right)}
       {\sum_{\ell}\exp\!\left(\boldsymbol{q}_i^\mathsf{T}
       \boldsymbol{\mathcal{T}}_{i\leftarrow \ell}\boldsymbol{k}_{\ell}/\sqrt{d_h}\right)}, \\
    \boldsymbol{o}_i
    &= \sum_j \alpha_{ij}\boldsymbol{\mathcal{T}}_{i\leftarrow j}\boldsymbol{v}_j.
\end{align}
Here $\boldsymbol{\mathcal{T}}_{i\leftarrow j}$ maps the key and value representations at position $j$ to the coordinate system of query position $i$.
The camera embeddings $\boldsymbol{r}$ then query $\boldsymbol{z}$ through $D_\theta$ to synthesize the target features $\hat{\boldsymbol{F}}_t$:
\begin{align}
    \boldsymbol{H}'        & = \text{Transformer}(\text{Concat}[\boldsymbol{r}, \boldsymbol{z}]), \\
    \hat{\boldsymbol{F}}_t & = \{\boldsymbol{H}'_i : i\in\mathcal{I}_t\},
\end{align}
where $\mathcal{I}_t=\{1,\ldots,N_T N_p\}$ indexes the $N_p$ patch tokens from each of the $N_T$ target views; the following $N_q$ outputs correspond to the latent tokens and are not passed to the prediction heads.

\paragraph{Losses}
The trainable Spa3R Encoder, Decoder, and prediction heads are optimized jointly, while VGGT, its camera head, and DINOv3 remain frozen. The decoder has separate geometric ($g$) and semantic ($s$) prediction heads. For head $h\in\{g,s\}$, the loss is computed independently for every target patch token and then averaged over target views and patches. Let $\hat{\boldsymbol{F}}^{(h)}_{t,p}$ and $\boldsymbol{F}^{(h)}_{t,p}$ denote the predicted and aligned target feature vectors at patch $p$ of view $t$. We use
\begin{align}
    \mathcal{L}_h
    = \frac{1}{N_T N_p}\sum_{t=1}^{N_T}\sum_{p=1}^{N_p}
    \Bigg[&\frac{1}{D_h}\left\|\hat{\boldsymbol{F}}^{(h)}_{t,p}-\boldsymbol{F}^{(h)}_{t,p}\right\|_1 \notag \\
    &+1-\frac{(\hat{\boldsymbol{F}}^{(h)}_{t,p})^\mathsf{T}\boldsymbol{F}^{(h)}_{t,p}}
    {\|\hat{\boldsymbol{F}}^{(h)}_{t,p}\|_2\|\boldsymbol{F}^{(h)}_{t,p}\|_2+\epsilon}\Bigg], \\
    \mathcal{L}_{\mathrm{PSFM}} &= \lambda_g\mathcal{L}_g+\lambda_s\mathcal{L}_s,
\end{align}
where $D_h$ is the feature dimension and $\epsilon$ is a small constant for numerical stability. We set $\lambda_g=\lambda_s=1$.

\subsection{Grounding Reasoning in Spatial Context}
\label{sec:SPA_VLM}

To provide VLMs with spatial context, we integrate the pre-trained, frozen Spa3R Encoder into Qwen2.5-VL~\cite{Qwen2_5VL} to create Spa3-VLM. As shown in \cref{fig:framework}(c), a lightweight Residual Cross-Attention Adapter fuses the unified, view-invariant Spa3R latent representation $\boldsymbol{z}$ with the VLM's native 2D visual features $\boldsymbol{F}_V$.
We first project $\boldsymbol{z}$ to the VLM hidden dimension and then use $\boldsymbol{F}_V$ as queries in cross-attention over the projected latent $\tilde{\boldsymbol{z}}$. The cross-attention output is normalized, transformed by a gated MLP, and added residually to $\boldsymbol{F}_V$:
\begin{align}
    \boldsymbol{F}_{\text{ctx}}   & = \text{CrossAttn}(q=\boldsymbol{F}_V, k=\boldsymbol{z}, v=\boldsymbol{z}), \\
    \boldsymbol{F}_V'             & = \boldsymbol{F}_V + \text{MLP}(\boldsymbol{F}_{\text{ctx}}).
\end{align}

The MLP's output projection is zero-initialized so that the adapter initially preserves the native visual features. The spatially enriched features $\boldsymbol{F}_V'$ are then concatenated with text embeddings and fed into the language model. The adapter and language model are fine-tuned through spatial instruction tuning, while the Spa3R Encoder and VLM vision encoder remain frozen. This design retains the base VLM's pre-trained knowledge while grounding its reasoning in global spatial context.

\section{Experiments}

We first describe the datasets and evaluation benchmarks in \cref{sec:setup}, followed by implementation details for Spa3R pre-training and Spa3-VLM instruction tuning in \cref{sec:implementation}. We report the main results in \cref{sec:main_results}, followed by ablations and qualitative analyses in \cref{sec:ablation,sec:visualization}.

\subsection{Setup}
\label{sec:setup}

\paragraph{Pre-training Datasets}
We pre-train Spa3R on the training splits of ScanNet~\cite{ScanNet} and ScanNet++~\cite{ScanNet++}, which comprise multi-view indoor RGB sequences captured in real-world environments. Only RGB frames are supplied to Spa3R; depth and ground-truth camera poses are not used as model inputs. No validation or test scenes from the pre-training datasets are used for optimization.

\paragraph{Evaluation Benchmarks}
We primarily evaluate on VSI-Bench~\cite{VSIBench}, a benchmark for visual-spatial reasoning over videos. It comprises over 5,000 question-answer pairs derived from 288 real-world indoor scene videos, categorized into configurational, measurement-estimation, and spatiotemporal tasks. We report accuracy for multiple-choice answer (MCA) tasks and mean relative accuracy (MRA) for numerical answer (NA) tasks. We additionally evaluate on CV-Bench~\cite{Cambrian1}, SPAR-Bench~\cite{SPAR}, and ViewSpatial-Bench~\cite{ViewSpatial} to cover complementary benchmark formulations and image/video input formats.

\definecolor{navyblue}{HTML}{0071BC}

\begin{table*}[t]
    \centering
    \newcolumntype{C}[1]{>{\centering\arraybackslash}m{#1}}
    \caption{Quantitative comparison on the VSI-Bench~\cite{VSIBench} spatial reasoning benchmark}
    \begin{tabular}{r|C{0.96cm}|C{0.96cm}C{0.96cm}C{0.96cm}C{0.96cm}C{0.96cm}C{0.96cm}C{0.96cm}C{0.96cm}}
        \toprule
                                                                           &                  &
        \rotatebox{55}{\scriptsize{\textbf{Obj. Count}}}                   &
        \rotatebox{55}{\scriptsize{\textbf{Abs. Dist.}}}                   &
        \rotatebox{55}{\scriptsize{\textbf{Obj. Size}}}                    &
        \rotatebox{55}{\scriptsize{\textbf{Room Size}}}                    &
        \rotatebox{55}{\scriptsize{\textbf{Rel. Dist.}}}                   &
        \rotatebox{55}{\scriptsize{\textbf{Rel. Dir.}}}                    &
        \rotatebox{55}{\scriptsize{\textbf{Route Plan}}}                   &
        \rotatebox{55}{\scriptsize{\textbf{Appr. Order}}}                                                                                                                                                                                          \\
        \textbf{Model}                                                     & \textbf{Avg.}    &
        \multicolumn{4}{c}{\cellcolor{orange!10}\textbf{Numerical Answer}} &
        \multicolumn{4}{c}{\cellcolor{yellow!10}\textbf{Multiple-Choice Answer}}                                                                                                                                                                   \\
        \midrule

        \rowcolor{navyblue!5}
        \multicolumn{1}{l|}{\textcolor{black}{\textit{Proprietary Models}}}
                                                                           &                  &                  &                  &                  &                  &                  &                  &               &                  \\
        GPT-4o~\cite{GPT4o}                                                & 34.0             & 46.2             & 5.3              & 43.8             & 38.2             & 37.0             & 41.3             & 31.5          & 28.5             \\
        Gemini-1.5-Flash~\cite{Gemini1_5}                                  & 42.1             & 49.8             & 30.8             & 53.5             & 54.4             & 37.7             & 41.0             & 31.5          & 37.8             \\
        Gemini-1.5-Pro~\cite{Gemini1_5}                                    & 45.4             & 56.2             & 30.9             & 64.1             & 43.6             & 51.3             & 46.3             & 36.0          & 34.6             \\
        \midrule

        \rowcolor{navyblue!5}
        \multicolumn{1}{l|}{\textcolor{black}{\textit{Open-source Models}}}
                                                                           &                  &                  &                  &                  &                  &                  &                  &               &                  \\
        InternVL2-8B~\cite{InternVL}                                       & 37.5             & 31.3             & 29.0             & 48.9             & 44.2             & 38.0             & 33.4             & 28.9          & 46.4             \\
        VILA-1.5-8B~\cite{VILA}                                            & 28.9             & 17.4             & 21.8             & 50.3             & 18.8             & 32.1             & 34.8             & 31.0          & 24.8             \\
        LLaVA-Video-7B~\cite{LLaVA-Video}                                  & 35.6             & 48.5             & 14.0             & 47.8             & 24.2             & 43.5             & 42.4             & 34.0          & 30.6             \\
        LLaVA-OneVision-7B~\cite{LLaVA-OV}                                 & 32.4             & 47.7             & 20.2             & 47.4             & 12.3             & 42.5             & 35.2             & 29.4          & 24.4             \\
        Qwen2.5-VL-3B~\cite{Qwen2_5VL}                                     & 30.6             & 24.3             & 24.7             & 31.7             & 22.6             & 38.3             & 41.6             & 26.3          & 21.2             \\
        Qwen2.5-VL-7B~\cite{Qwen2_5VL}                                     & 33.0             & 40.9             & 14.8             & 43.4             & 10.7             & 38.6             & 38.5             & 33.0          & 29.8             \\
        \midrule

        \rowcolor{navyblue!5}
        \multicolumn{1}{l|}{\textcolor{black}{\textit{Spatial Models}}}
                                                                           &                  &                  &                  &                  &                  &                  &                  &               &                  \\
        Spatial-MLLM-4B~\cite{SpatialMLLM}                                 & 48.4             & 65.3             & 34.8             & 63.1             & 45.1             & 41.3             & 46.2             & 33.5          & 46.3             \\
        VG-LLM-4B~\cite{VGLLM}                                             & 47.3             & 66.0             & 37.8             & 55.2             & 59.2             & 44.6             & 45.6             & 33.5          & 36.4             \\
        VG-LLM-8B~\cite{VGLLM}                                             & 50.7             & 67.9             & 37.7             & 58.6             & \underline{62.0} & 46.6             & 40.7             & 32.4          & 59.2             \\
        Cambrian-S-3B~\cite{CambrianS}                                     & \underline{57.3} & \textbf{70.7}    & \textbf{40.6}    & \underline{68.0} & 46.3             & \textbf{64.8}    & \textbf{61.9}    & 27.3          & \textbf{78.8}    \\
        \textbf{Spa3-VLM-4B \footnotesize{(Ours)}}                         & \textbf{58.6}    & \underline{69.0} & \underline{39.3} & \textbf{70.6}    & \textbf{62.8}    & \underline{57.9} & \underline{59.5} & \textbf{36.1} & \underline{73.6} \\
        \bottomrule
    \end{tabular}
    \label{tab:vsibench}
\end{table*}

\begin{table*}[t]
    \centering
    \newcolumntype{C}[1]{>{\centering\arraybackslash}m{#1}}
    \caption{Quantitative comparison across CV-Bench~\cite{Cambrian1}, SPAR-Bench~\cite{SPAR}, and ViewSpatial-Bench~\cite{ViewSpatial}}
    \begin{tabular}{lC{1cm}C{1cm}C{1cm}C{1.6cm}C{1.6cm}}
        \toprule
        \multirow{2}{*}{\textbf{Model}}                            & \multicolumn{3}{c}{\textbf{CV}} & \multirow{2}{*}{\textbf{SPAR}} & \multirow{2}{*}{\textbf{ViewSpatial}}                                       \\
        \cmidrule(lr){2-4}
                                                                   & 2D                              & 3D                             & Avg.                                  &                  &                  \\
        \midrule
        Qwen2.5-VL-3B~\footnotesize{\cite{Qwen2_5VL}}              & 69.1                            & 72.2                           & 70.6                                  & 24.6             & 35.6             \\
        SpatialLadder-3B~\footnotesize{\cite{SpatialLadder}}       & \underline{72.4}                & 74.9                           & 73.7                                  & 34.4             & \textbf{44.2}    \\
        Spatial-MLLM-4B~\footnotesize{\cite{SpatialMLLM}}          & -                               & -                              & \underline{73.8}                      & 35.1             & 43.6             \\
        SpatialThinker-3B~\footnotesize{\cite{SpatialThinker}}     & 71.0                            & \underline{76.3}               & 73.6                                  & -                & -                \\
        SpaceR-7B~\cite{SpaceR}                                    & -                               & -                              & -                                     & \underline{37.6} & 37.3             \\
        \rowcolor{gray!10} \textbf{Spa3-VLM-4B \footnotesize{(Ours)}} & \textbf{72.9}                   & \textbf{78.3}                  & \textbf{75.6}                         & \textbf{58.4}    & \underline{43.9} \\
        \bottomrule
    \end{tabular}
    \label{tab:multi_bench}
\end{table*}

\paragraph{Instruction-Tuning Datasets}
For the video-centric VSI-Bench, we fine-tune Spa3-VLM on the VSI-590K~\cite{CambrianS} dataset. For image-based benchmarks, we construct a composite training set comprising SPAR-234K~\cite{SPAR}, LLaVA-Hound~\cite{LLaVA-Video}, and VLM3R~\cite{VLM3R}, following prior work.

\subsection{Implementation Details}
\label{sec:implementation}

\paragraph{Spa3R Pre-training}
We initialize the Asymmetric View Aggregator and camera head with pre-trained weights from VGGT~\cite{VGGT}, which remain frozen during training. The geometric targets are the final-layer patch tokens from the VGGT aggregator. For semantic targets, we use the normalized final-layer patch tokens from DINOv3 ViT-L/16~\cite{DINOv3}. DINOv3 inputs are resized to $224\times224$ and normalized with ImageNet statistics, and its outputs are bilinearly aligned to the VGGT patch grid. The Spa3R Encoder and Decoder are implemented as 6-layer Transformers with a hidden dimension of $D=768$, 12 attention heads, a feed-forward expansion ratio of 4, and four register tokens. We use $N_q=256$ learnable query embeddings to form the spatial latent representation.

Each RGB frame is resized to a width of 518 pixels while preserving its aspect ratio and then center-cropped to dimensions divisible by 14. In each iteration, we sample 4 to 12 views from a scene and randomly designate half as context views and the remainder as target views. A dynamic batch sampler constructs batches of up to 24 images per GPU. Spa3R is pre-trained for 80K steps on 8 NVIDIA GeForce RTX 5090 GPUs using FP16 mixed precision with gradient accumulation over two steps. We use AdamW with a learning rate of $2 \times 10^{-3}$, $(\beta_1,\beta_2)=(0.9,0.999)$, and a weight decay of $10^{-4}$. The learning rate follows a cosine schedule from $2 \times 10^{-3}$ to $2 \times 10^{-4}$ over 80K steps without warmup.

\paragraph{Spa3-VLM Instruction Tuning}
We adopt Qwen2.5-VL-3B~\cite{Qwen2_5VL} as the base VLM. During instruction tuning, the pre-trained Spa3R Encoder and the VLM's native vision encoder remain frozen to preserve their learned representations. The adapter uses a two-layer projector to map the 768-dimensional Spa3R latents to the 2,048-dimensional VLM space, followed by a 16-head cross-attention layer, LayerNorm, and a gated two-layer MLP with a 512-dimensional bottleneck. The MLP's output layer is zero-initialized before residual fusion. We fine-tune the adapter and language model for 1 epoch using BF16 and DeepSpeed ZeRO-2. With eight GPUs, the per-device batch size is 1 and gradients are accumulated over eight steps, giving an effective batch size of 64. Both components use AdamW with a learning rate of $10^{-5}$, $(\beta_1,\beta_2)=(0.9,0.999)$, and a weight decay of 0.01. The learning rate follows a cosine schedule with 3\% warmup, and the gradient norm is clipped to 1. For video instruction tuning, we uniformly sample 4 to 8 frames per video. The geometry branch resizes each frame to a width of 518 pixels, while the native VLM branch uses a maximum pixel budget of $576\times28^2$ per frame. During evaluation, we uniformly sample 32 frames per video and use greedy decoding with temperature 0, one beam, and at most 4,096 newly generated tokens.

\subsection{Main Results}
\label{sec:main_results}

Quantitative results are reported in \cref{tab:vsibench,tab:multi_bench}. On VSI-Bench, Spa3-VLM obtains an average score of 58.6\%, 1.3 points above Cambrian-S-3B (57.3\%). It ranks first among the compared methods on object size (70.6\%), room size (62.8\%), and route planning (36.1\%), while Cambrian-S remains stronger on relative distance and appearance order.

Across the image-based evaluations, Spa3-VLM obtains 75.6\% on CV-Bench and 58.4\% on SPAR-Bench, outperforming the best previously reported results listed in \cref{tab:multi_bench} by 1.8 and 20.8 points, respectively. On ViewSpatial-Bench, it reaches 43.9\%, 0.3 points below SpatialLadder-3B. These results indicate that the gains extend beyond the video-centric setting.

\subsection{Novel-View Spatial Field Prediction}
\label{sec:novel_view_prediction}

\begin{table}[t]
    \centering
    \caption{Novel-view spatial field prediction on ScanNet++}
    \begin{tabular}{ll}
        \toprule
        \textbf{Metric} & \textbf{Score} \\
        \midrule
        \rowcolor{gray!15}
        \multicolumn{2}{l}{\textbf{VGGT geometry feature}} \\
        L1 $\downarrow$     & 0.5280 $\pm$ 0.1398 \\
        Cosine $\uparrow$   & 0.8806 $\pm$ 0.1200 \\
        \midrule
        \rowcolor{gray!15}
        \multicolumn{2}{l}{\textbf{DINOv3 semantic feature}} \\
        L1 $\downarrow$     & 0.1371 $\pm$ 0.0119 \\
        Cosine $\uparrow$   & 0.7106 $\pm$ 0.0602 \\
        \midrule
        \rowcolor{gray!15}
        \multicolumn{2}{l}{\textbf{Depth}} \\
        AbsRel $\downarrow$ & 0.2492 $\pm$ 0.0724 \\
        RMSE $\downarrow$   & 0.5734 $\pm$ 0.2234 m \\
        $\delta_1 \uparrow$ & 0.6434 $\pm$ 0.1138 \\
        \midrule
        \rowcolor{gray!15}
        \multicolumn{2}{l}{\textbf{Aggregated point cloud}} \\
        Chamfer-L1 $\downarrow$ & 0.1583 $\pm$ 0.0464 m \\
        F-score@5 cm $\uparrow$  & 0.2541 $\pm$ 0.0573 \\
        F-score@10 cm $\uparrow$ & 0.5001 $\pm$ 0.0766 \\
        \bottomrule
    \end{tabular}
    \label{tab:novel_view_prediction}
\end{table}

\Cref{tab:novel_view_prediction} evaluates the pre-trained spatial representation independently of language reasoning. In this evaluation, Spa3R receives eight context views and predicts feature fields at novel views along a dense novel-view trajectory. The predicted fields achieve cosine similarities of 0.8806 and 0.7106 for VGGT and DINOv3 features, respectively, indicating that the latent supports the recovery of both geometric and semantic targets beyond the observed views.

To probe the scale-aligned geometry encoded in these predictions, we fit a lightweight depth head while keeping Spa3R frozen. With a single scale alignment per scene, the recovered depth achieves an AbsRel of 0.2492, an RMSE of 0.5734 m, and a $\delta_1$ accuracy of 0.6434 against sensor depth. We then back-project and aggregate the target-view depth predictions into a point cloud, yielding a Chamfer-L1 of 0.1583 m and F-scores of 0.2541 and 0.5001 at 5 cm and 10 cm, respectively, against the geometry observable along the target trajectory. The higher F-score at 10 cm than at 5 cm indicates that the representation captures coarse scene structure more reliably than fine surface detail. Because the encoder observes only the context views, these results further suggest that the latent geometry generalizes to novel viewpoints and to regions occluded in individual context images. Overall, the evaluation indicates that novel-view geometry can be recovered from the learned representation.

\subsection{Ablation Studies}
\label{sec:ablation}

We conduct ablation studies on VSI-Bench to assess the core design choices of our framework.

\paragraph{Synergy of Geometric and Semantic Prediction Targets}
We analyze the impact of the supervision signals used as Spa3R pre-training targets. \Cref{tab:ablat_tgt_feats} compares geometric features from VGGT, semantic visual features from DINOv3, vision-language features from CLIP, and combinations thereof. VGGT alone reaches 57.5\%, DINOv3 alone reaches 56.7\%, and CLIP alone reaches 51.9\%. Combining VGGT and DINOv3 yields the highest average of 58.6\%, suggesting that the two feature spaces provide complementary signals for geometric structure and high-level semantic context.

\begin{table}[ht]
    \centering
    \caption{Ablation of feature-prediction targets for Spa3R pre-training}
    \begin{tabular}{cccccc}
        \toprule
        \textbf{VGGT}                 & \textbf{DINOv3} & \textbf{CLIP} & \textbf{NA (MRA)} & \textbf{MCA (Acc.)} & \textbf{Avg.} \\
        \midrule
        \checkmark                    &               &               & \textbf{60.5}      & 54.4                  & 57.5          \\
                                      & \checkmark    &               & 59.0               & 54.4                  & 56.7          \\
                                      &               & \checkmark    & 59.5               & 44.3                  & 51.9          \\
        \rowcolor{gray!10} \checkmark & \checkmark    &               & 60.4               & \textbf{56.8}         & \textbf{58.6} \\
        \bottomrule
    \end{tabular}
    \label{tab:ablat_tgt_feats}
\end{table}

\paragraph{Effectiveness of Unified Spatial Representation}
To assess the benefit of PSFM, we compare Spa3R against a baseline that directly supplies view-conditioned VGGT features to the VLM, which is architecturally equivalent to VG-LLM~\cite{VGLLM}. As shown in \cref{tab:ablat_spa_repr}, direct VGGT features improve the matched Qwen2.5-VL-3B fine-tuning baseline from 50.9\% to 55.1\%, while Spa3R reaches 58.6\%, a further gain of 3.5 points. This controlled comparison supports the utility of compressing multi-view features through the PSFM objective and isolates the contribution of the spatial branch under matched downstream training.

\begin{table}[ht]
    \centering
    \caption{Ablation on spatial representation paradigms}
    \begin{tabular}{lccc}
        \toprule
        \textbf{Spatial Repr.}             & \textbf{NA (MRA)} & \textbf{MCA (Acc.)} & \textbf{Avg.} \\
        \midrule
        None                              & 58.4               & 43.5                  & 50.9          \\
        VGGT                              & \textbf{60.5}      & 49.5                  & 55.1          \\
        \rowcolor{gray!10} \textbf{Spa3R} & 60.4               & \textbf{56.8}         & \textbf{58.6} \\
        \bottomrule
    \end{tabular}
    \label{tab:ablat_spa_repr}
\end{table}

\begin{figure*}
    \includegraphics[width=\textwidth]{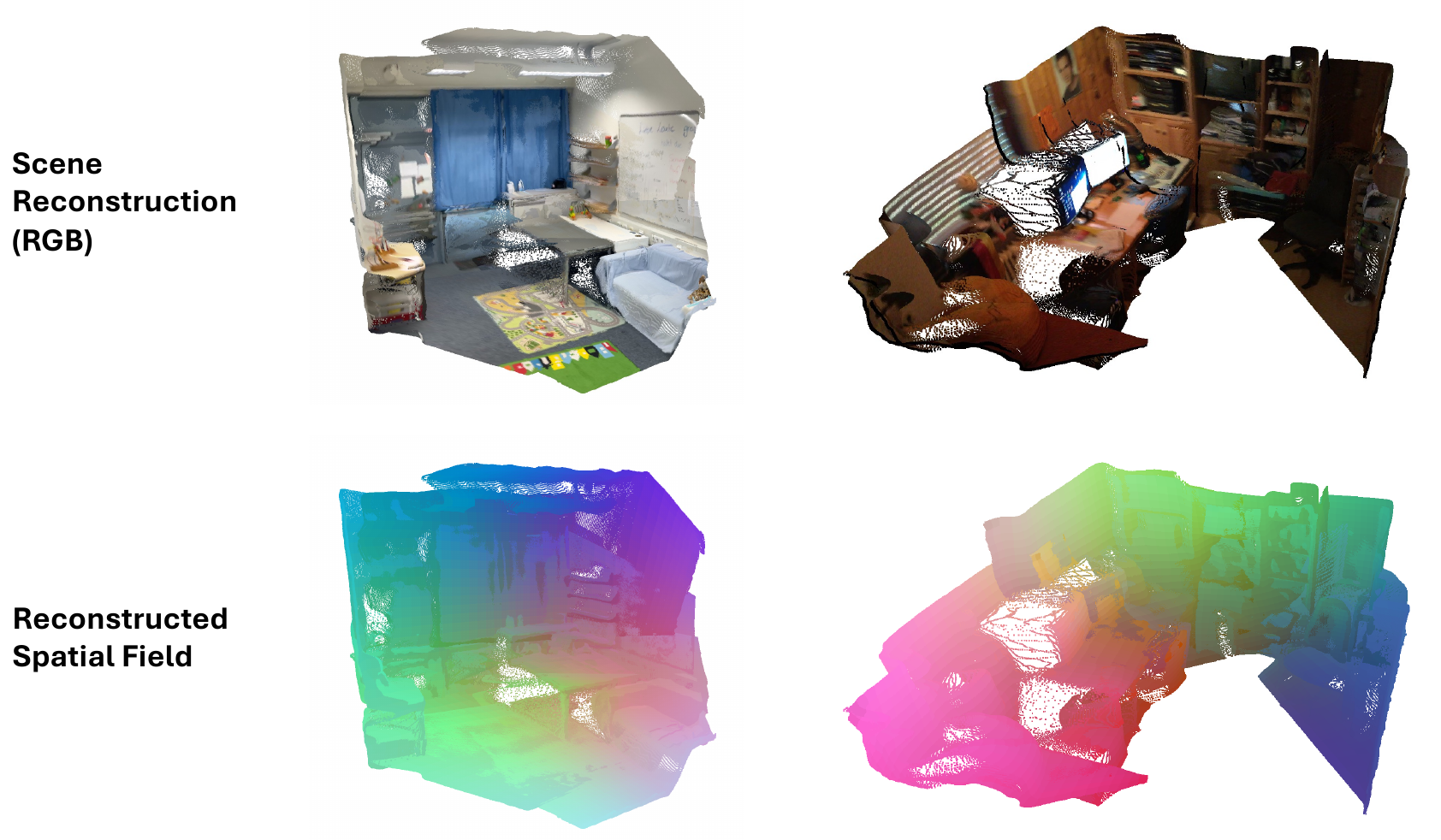}
    \caption{Qualitative visualization of learned feature fields in 3D space. After unprojection onto the VGGT point cloud, features from different views show consistent spatial organization and smooth transitions. These examples provide qualitative evidence that the predicted features preserve the underlying scene geometry across views.}
    \label{fig:visualization_3d}
\end{figure*}

\paragraph{Architecture for Spa3-VLM Integration}
We further compare ways to integrate the spatial latent into the VLM. As reported in \cref{tab:ablat_vlm_intgr}, sequence appending reaches 51.1\%, only 0.2 points above the matched baseline, whereas residual cross-attention reaches 58.6\%. We hypothesize that, with sequence appending, the pre-trained VLM preferentially attends to its native visual tokens and underutilizes the newly introduced spatial tokens. By contrast, cross-attention lets each native visual token retrieve relevant spatial context before the residual update.

\begin{table}[th]
    \centering
    \caption{Ablation on VLM integration architectures}
    \begin{tabular}{lccc}
        \toprule
        \textbf{Integration}           & \textbf{NA (MRA)} & \textbf{MCA (Acc.)} & \textbf{Avg.} \\
        \midrule
        None                           & 58.4               & 43.5                  & 50.9          \\
        Seq. Append.                   & 58.5               & 43.7                  & 51.1          \\
        \rowcolor{gray!10} Cross-Attn. & \textbf{60.4}      & \textbf{56.8}         & \textbf{58.6} \\
        \bottomrule
    \end{tabular}
    \label{tab:ablat_vlm_intgr}
\end{table}

\paragraph{PSFM Target-View Ratio}
We ablate the proportion of views assigned to the target set, which controls the amount of available context and the number of prediction targets. As shown in \cref{tab:ablat_mask_ratio}, the 50\% split gives the highest average score (58.6\%), compared with 57.5\% at 25\% and 58.1\% at 75\%. This result is consistent with a trade-off between context coverage and prediction difficulty.

\begin{table}[ht]
    \centering
    \caption{Ablation of the target-view ratio used for PSFM pre-training}
    \begin{tabular}{cccc}
        \toprule
        \textbf{Mask ratio}     & \textbf{NA (MRA)} & \textbf{MCA (Acc.)} & \textbf{Avg.} \\
        \midrule
        25\%                    & 58.9               & 56.1                  & 57.5          \\
        \rowcolor{gray!10} 50\% & \textbf{60.4}      & \textbf{56.8}         & \textbf{58.6} \\
        75\%                    & 59.6               & 56.7                  & 58.1          \\
        \bottomrule
    \end{tabular}
    \label{tab:ablat_mask_ratio}
\end{table}

\paragraph{Camera Embedding Mechanism}
We investigate the mechanism for conditioning the decoder on the target viewpoint. As shown in \cref{tab:ablat_cam_emb}, PRoPE~\cite{PRoPE} improves the average score from 57.6\% with Pl\"ucker coordinates~\cite{Plucker} to 58.6\%. PRoPE represents camera intrinsics, extrinsics, and image-plane patch positions through relative attention operators, whereas the compared Pl\"ucker embedding is injected at the token level.

\begin{figure*}
    \includegraphics[width=0.95\textwidth]{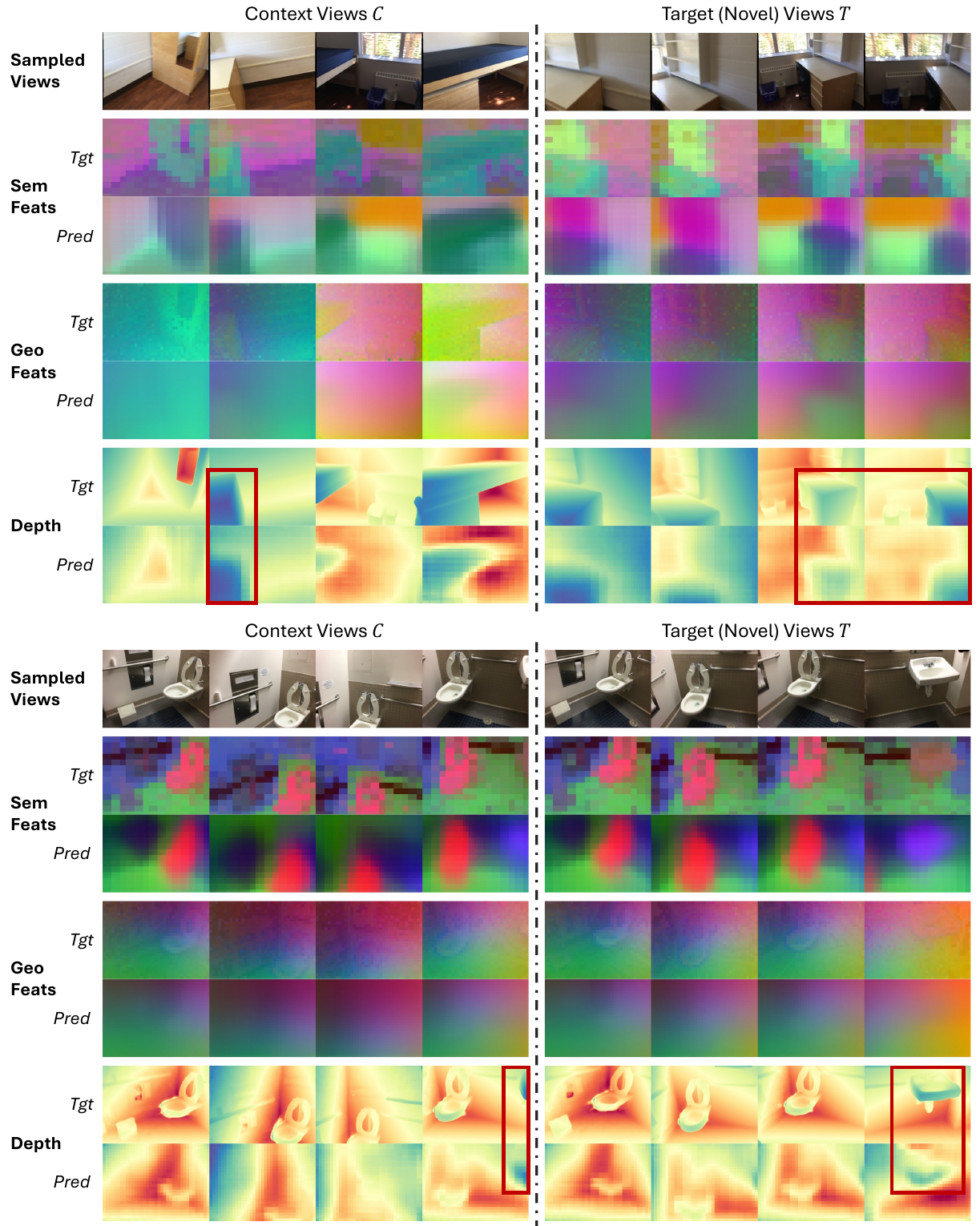}
    \caption{Qualitative visualization of learned feature fields in novel views. Predicted features show continuous and spatially coherent layouts similar to the aligned targets. Colors do not match exactly because separate PCA projections are fitted to the two feature sets. The red boxes highlight predictions in regions that are occluded or unobserved in the context; these examples provide qualitative evidence of cross-view consistency within the sampled target views.}
    \label{fig:visualization}
\end{figure*}

\begin{table}[ht]
    \centering
    \caption{Ablation of target-view encoding in the Spa3R Decoder}
    \begin{tabular}{lccc}
        \toprule
        \textbf{Camera Enc.} & \textbf{NA (MRA)} & \textbf{MCA (Acc.)} & \textbf{Avg.} \\
        \midrule
        Plücker                  & 59.9               & 55.4                  & 57.6          \\
        \rowcolor{gray!10} PRoPE & \textbf{60.4}      & \textbf{56.8}         & \textbf{58.6} \\
        \bottomrule
    \end{tabular}
    \label{tab:ablat_cam_emb}
\end{table}

\subsection{Qualitative Analysis}
\label{sec:visualization}

We visualize the reconstructed feature fields in \cref{fig:visualization_3d} to examine the representational capacity of Spa3R. Specifically, we project the original RGB images and predicted feature maps from all views onto the canonical point-cloud geometry reconstructed by VGGT~\cite{VGGT}. In \cref{fig:visualization}, we sample eight views from each of two scenes and compare the target geometric and semantic features with the predictions synthesized by the Spa3R Decoder.

We use principal component analysis (PCA) to project the high-dimensional features into RGB space. We also train a separate MLP depth probe to regress depth maps from the predicted features $\hat{\boldsymbol{F}}_t$. Gradients from this probe are not back-propagated to Spa3R; it serves only as a diagnostic of the geometry encoded in the features.

As shown in \cref{fig:visualization_3d,fig:visualization}, the predicted features display spatial organization consistent with the corresponding target features across views. The PCA projections show continuous, coherent region-level layouts, and the detached depth probe recovers scene structure visually aligned with the VGGT reconstruction. Predictions in the highlighted occluded or unobserved regions further suggest that the decoder uses information aggregated across context views. The quantitative evaluation in \cref{sec:novel_view_prediction} complements these examples through comparison with sensor-derived geometry.

\section{Conclusion}

We introduced Spa3R, a self-supervised framework that compresses unposed multi-view RGB observations into a unified, view-invariant spatial representation by predicting geometric and semantic feature fields at novel target views. A residual cross-attention adapter integrates this representation into Qwen2.5-VL, yielding Spa3-VLM. Spa3-VLM reaches an average score of 58.6\% on VSI-Bench and improves over the matched fine-tuning baseline by 7.7 points. Controlled ablations further show gains from joint geometric--semantic targets, the PSFM representation, and cross-attention integration.

The present study focuses on static indoor scenes and inherits 3D priors and pose estimates from a frozen geometry foundation model. Although the experiments demonstrate gains in downstream spatial reasoning, the method does not yet model calibrated uncertainty, establish formal disentanglement, or demonstrate complete recovery of unobserved scene geometry. Extending predictive spatial representation learning to dynamic and outdoor environments is an important direction for future work.

Code is available at \url{https://github.com/hustvl/Spa3R}.

\appendix[Additional Qualitative Results and Resource Requirements]

\subsection{Parameter Counts and Inference Memory}
\label{app:efficiency}

Across all training and evaluation stages, the complete stack contains 5.381B unique parameters. Spa3R pre-training optimizes 103.4M of 1.597B loaded parameters, whereas Spa3-VLM instruction tuning optimizes 28.9M adapter parameters and 3.086B language-model parameters, totaling 3.115B trainable parameters. The Spa3-VLM inference path uses 4.737B unique parameters. DINOv3, the VGGT prediction heads, and the Spa3R Decoder and prediction heads are used only during pre-training or representation-level evaluation and are excluded from downstream inference. When processing 32 frames, Spa3-VLM requires 17.584 GiB of peak allocated memory.

\subsection{Qualitative Spa3-VLM Predictions}
\label{app:qualitative_predictions}

\begin{figure*}[t]
    \centering
    \includegraphics[width=0.85\textwidth]{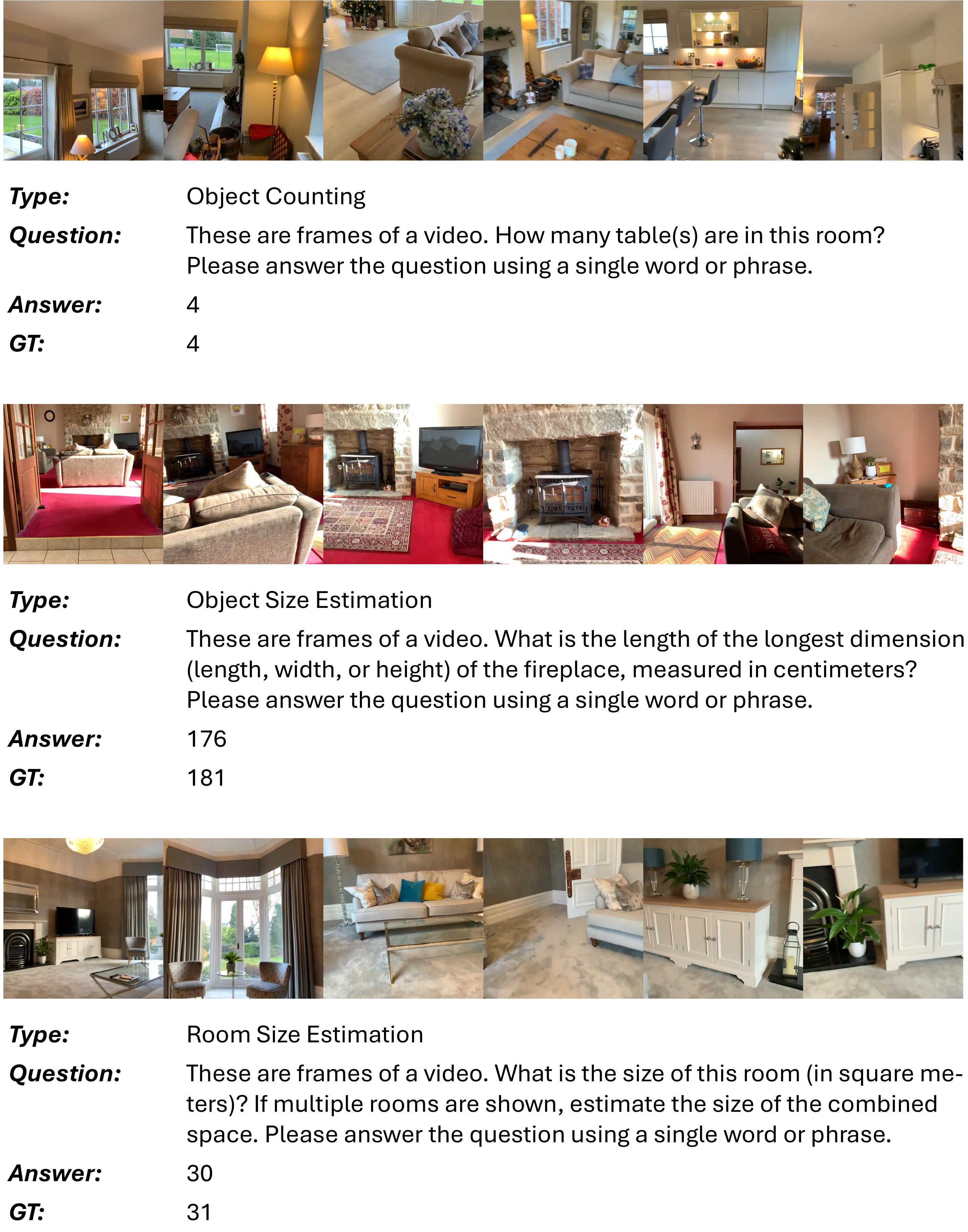}
    \caption{Qualitative Spa3-VLM predictions on VSI-Bench. Each example shows six representative frames sampled from the 32-frame input used by the evaluation protocol, followed by the task type, question, Spa3-VLM answer, and ground-truth answer. Spa3-VLM correctly counts repeated object instances and estimates object scale.}
    \label{fig:predictions}
\end{figure*}

\Cref{fig:predictions} presents selected downstream examples from VSI-Bench~\cite{VSIBench}. In these cases, Spa3-VLM aggregates observations across frames to count repeated object instances and estimate both object and room scale.

\clearpage
\bibliographystyle{IEEEtran}
\bibliography{main}

\end{document}